\documentclass[final,authoryear,5p,times,twocolumn]{elsarticle}

\usepackage{amsmath}
\usepackage{amssymb} % for real number R:

\usepackage{tikz}
\usetikzlibrary{arrows,matrix,positioning,patterns}
\usetikzlibrary{calc}
\usepackage{pgfplots} % to include tikz-figs.
\usetikzlibrary{external}
\usetikzlibrary{angles}
\usepackage{hyperref}%<----UNCOMMENT IF WANT THE LINKS !
\usetikzlibrary{plotmarks}

\newdefinition{rmk}{Remark}
\newproof{pf}{Proof}
\newproof{pot}{Proof of Theorem \ref{thm2}}

\usepackage[ruled, vlined, linesnumbered]{algorithm2e}

\usepackage{amsthm}

\newtheorem{problem}{Problem}

\usepackage{amsmath}
\usepackage{amssymb} % for real number R:

\usepackage{eurosym}
\usepackage{colortbl}

\usetikzlibrary{arrows.meta,calc,decorations.markings,math,arrows.meta}

\usepackage[labelformat=simple]{subcaption}
\DeclareCaptionLabelSeparator{periodspace}{.\quad}
\usepackage{color,framed}
\usepackage[utf8]{inputenc} % unter Windows latin1 statt utf8
\usepackage{tabularx}

\usepackage{rotating}
\usepackage{graphicx}
\usepackage{lscape}
\allowdisplaybreaks % for long eq. over multiple pages.
\usepackage{longtable}

\usepackage[utf8]{inputenc}
\usepackage{pgfplots}
\usepgfplotslibrary{groupplots} 
\pgfplotsset{compat=1.3}

\makeatletter
\def\ps@pprintTitle{%
 \let\@oddhead\@empty
 \let\@evenhead\@empty
 \def\@oddfoot{}%
 \let\@evenfoot\@oddfoot}
\makeatother
\begin{document}

\begin{frontmatter}
%
%
%---------
%%JOURNAL.
\title{Accelerated Sub-Image Search For Variable-Size Patches Identification\\Based On Virtual Time Series Transformation And Segmentation}
\author{Mogens Plessen\corref{cor1}}
\cortext[cor1]{MP is with Findklein GmbH, Switzerland, \texttt{mgplessen@gmail.com}}

%%%%%%%%%%%%%%%%%%%%%%%%%%%%%%%%%%%%%%%%%%%%%%%%%%%
\begin{abstract}
This paper addresses two tasks: (i) fixed-size objects such as hay bales are to be identified in an aerial image for a given reference image of the object, and (ii) variable-size patches such as areas on fields requiring spot spraying or other handling are to be identified in an image for a given small-scale reference image. Both tasks are related. The second differs in that identified sub-images similar to the reference image are further clustered before patches contours are determined by solving a traveling salesman problem. Both tasks are complex in that the exact number of similar sub-images is not known a priori. The main discussion of this paper is presentation of an acceleration mechanism for sub-image search that is based on a transformation of an image to multivariate time series along the RGB-channels and subsequent segmentation to reduce the 2D search space in the image. Two variations of the acceleration mechanism are compared to exhaustive search on diverse synthetic and real-world images. Quantitatively, proposed method results in solve time reductions of up to 2 orders of magnitude, while qualitatively delivering comparative visual results. Proposed method is neural network-free and does not use any image pre-processing.
\end{abstract}
%%%%%%%%%%%%%%%%%%%%%%%%%%%%%%%%%%%%%%%%%%%%%%%%%%%s
\begin{keyword}
Sub-Image Search; Aerial Images; Time Series; Variable-Size Patches; Spot Spraying.%; Agricultural logistics.
\end{keyword}
%%%%%%%%%%%%%%%%%%%%%%%%%%%%%%%%%%%%%%%%%%%%%%%%%%%
\end{frontmatter}

%%%%%%%%%%%%%%%%%%%%%%%%%%%%%%%%%%%%%%%%%%%%%%%%%
\section{Introduction\label{sec_intro}}

\begin{figure*}
\centering
  \includegraphics[width=.99\linewidth]{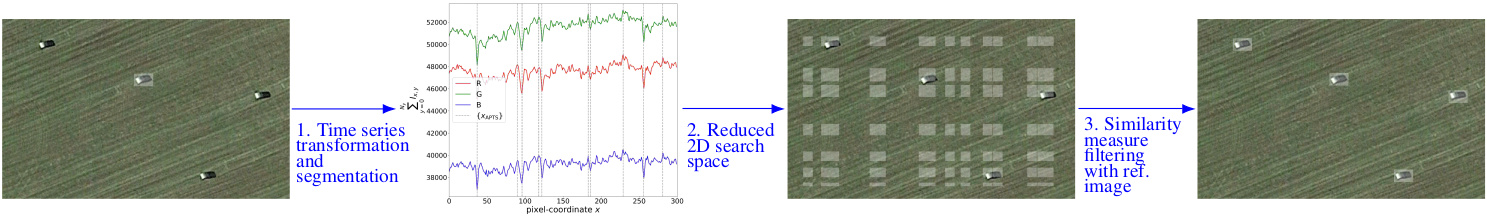}
\caption{Problem 1: Graphical abstract. \emph{Fixed-size} objects such as hay bales are to be identified in an image for a given smaller-scale reference image. Proposed hierarchical solution approach leverages a virtual time series transformation and segmentation to reduce the search space for sub-image search using similarity with respect to the reference image as filtering criterion. Note that the reduced search space shows the $(x,y)$-coordinates of the top-left corner of any sub-image measured in similarity to the reference image.}
\label{fig_prob1}
\end{figure*}

\begin{figure*}
\centering
  \includegraphics[width=.99\linewidth]{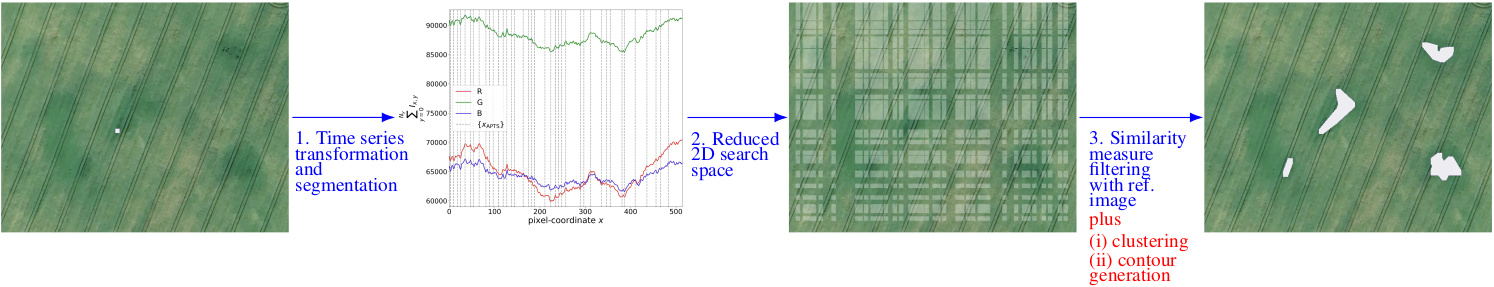}
\caption{Problem 2: Graphical abstract. \emph{Variable-size} patches such as areas on fields requiring spraying or other handling are to be identified in an image for a given smaller-scale reference patch. The approach to address this problem is identical to the approach for Problem \ref{problem1}, however with the extension that identified sub-images are further clustered and contours are generated to produce independent non-overlapping patches areas.}
\label{fig_prob2}
\end{figure*}

Using satellite images or other aerial images as a basis for analysis and decision taking is a trending topic. In agriculture, the growing usage of unmanned aerial vehicles (UAVs) is reinforcing this trend. This paper is motivated by the need to identify in aerial images of agricultural fields variable-size patches such as damaged crops or similar areas, before sending the contours of these patches to a path planning module for spot spraying with UAVs (\cite{plessen2024path}, \cite{li2023coverage}). 

It can be distinguished between two main image analysis classes for agricultural applications: (i) analysis of images taken close to the ground, and (ii) remote sensing applications with aerial images taken by satellite or aerial vehicles.
The former class includes e.g. weed detection (\cite{ahmad1999weed}, \cite{tang2016weed}) and crop-row identification (\cite{sogaard2003determination}, \cite{ospina2019simultaneous}). The latter class also includes weed detection applications but from a large distance and for larger areas (\cite{barrero2016weed}, \cite{islam2021early},  \cite{veeranampalayam2020comparison}, \cite{ferro2024comparison}, \cite{reedha2022transformer}).

% NOTATION.  
\begin{table}
\centering
\begin{tabular}{|ll|}
\hline
\multicolumn{2}{|c|}{MAIN NOMENCLATURE}\\
\multicolumn{2}{|l|}{Symbols}\\
$\mathbb{I}_{[0,255]}$ & Set of integers between 0 and 255, (-).\\
$\mathcal{X}_\text{space}$ & Image search space along rows, (-).\\
$\mathcal{Y}_\text{space}$ & Image search space along columns, (-).\\
$\Delta_x,\Delta_y$ & Stride step-size for sub-image search, (-).\\
$I$ & Input image of dimension $N_x \times N_y$, (-).\\
$i^\text{ref}$ & Reference image of dimension $N_x^\text{ref} \times N_y^\text{ref}$, (-).\\
$\tau_\text{solve}$ & Solve time, (s).\\[3pt]
\multicolumn{2}{|l|}{Hyperparameters}\\
$p$ & Tolerance for search space reduction, (-).\\
$M$ & Upper bound on sub-images, (-).\\
$K^\text{max}$ & APTS bound on segmentation instances, (-).\\
$\epsilon^\text{min}$, $\epsilon^\text{max}$ & APTS transaction cost level bounds, (-).\\
$\gamma^\text{mult}$, $\gamma^\text{close}$ & APTS multiplier and tolerance, (-).\\[3pt]
\multicolumn{2}{|l|}{Functions}\\
$\mathcal{F}^\text{APTS}(\cdot)$ & Calculating segmentation instances via APTS.\\
$\mathcal{F}^\text{cost}(\cdot)$ & Calculating cost during sub-image search.\\
$\mathcal{F}^{j_\text{axis}}(\cdot)$ & Calculating an approximation of a sub-image.\\[3pt]
\multicolumn{2}{|l|}{Abbreviations}\\
APTS & A Posteriori Trading-inspired Segmentation.\\
TSP & Traveling Salesman Problem.\\
UAV & Unmanned Aerial Vehicle.\\
\hline
\end{tabular}
%\vspace{-0.4cm}
\end{table} 

Many recent research on image analysis in agriculture focused on deep learning based methods. However, this approach has multiple disadvantages. Besides the difficulty of network architecture choice and hardware setup, one crucial practical disadvantage is that \emph{labeled} training data is required. In \cite{reedha2022transformer} this is acknowledged as '\emph{manual image labeling being a very time consuming task which implying huge labor costs}'. Furthermore, labeling data may be subjective. For practical variable-size patches identification this is disadvantageous. For example, one farmer or user may deem a specific patch area as requiring spot spraying, while another more cost-conscious farmer might not deem it as needing it. For these scenarios it is more desirable to have a hyperparameter choice that quickly permits to cater to the user's preference. For labeling- and deep learning based methods this is not easily achieved, since the neural network would have to be retrained. For these reasons, in this paper a neural network-free approach is adopted. A reference image in combination with a similarity measure and hyperparameter choice for user preference is used. 

The general solution approach taken in this paper can be classified as \emph{sub-image search} (\cite{sebe1999multi}, \cite{ke2004efficient}). The problems addressed in this paper are related to the \emph{Location Problem} in \cite{sebe1999multi}, however, with an important difference. Instead of '\textit{finding for a given query image $\mathcal{Q}$ and an image $\mathcal{I}$ the location in $\mathcal{I}$ where $\mathcal{Q}$ is present}', here \emph{all} locations of sub-images that are similar to the query image according to a similarity measure shall be found. 

The main objective of this paper is to present a method to \emph{accelerate} sub-image search. This is achieved by two steps: (i) transformation of an image to virtual multivariate time series along the RGB-channels for two separate cases along row- and column-dimensions of the image, before (ii) segmenting the time series and using the segmentation points as sampling points for search space reduction in the 2D image space. In \cite{eom1990recognition}, \cite{zhang2004review} 1D \emph{centroidal profiles} are discussed to describe shapes in images. Similarly, in \cite{keogh2006lb_keogh} 2D shapes are converted to 1D pseudo time series by calculating the distance from every point on a shape contour to the central point and treating it as the vertical-axis of a time series. For our case this method of time series generation is unsuitable for two reasons. First, shape contour can here not be employed for time series generation since for patches identification and sub-image search in an agricultural setting there typically is no shape contour available that could be used. Second, for centroidal profiles it is not straightforward to keep track of spatial 2D information useful for sub-image search. This is since typically the horizontal-axis of a centroidal profile generated time series would map non-linearly to the row and column space of an image (a similar argument can be constructed to also exclude \emph{color histograms} which also are sometimes used to create time series from images). For these two reasons, another time series generation scheme is used. The image is separately summed along row and columns to generate separate multivariate time series along the RGB-channels. This permits to directly map time series indices back to the 2D image space along rows and columns for sub-image search.

Aforementioned segmentation of the generated time series is carried out by the APTS-algorithm (A Posteriori Trading-inspired Segmentation, \cite{plessen2023posteriori}), which was shown to perform faster compared to alternative methods. Segmenting time series is useful for many applications. For example, in \cite{plessen2020integrated} it is used to generate an alphabet of shapelets based on the segmentation of centroids for clustered data. In the present paper, segmentation instances of the time series are used as basic sampling points for 2D search space reduction for sub-image search.

The research gap and motivation for this paper is discussed. There is a research gap in the application of time series analysis algorithms to image analysis tasks after the transformation of images to virtual time series. Within the agricultural context, this is the first paper that proposes such an approach for sub-image search, building upon a recent segmentation algorithm.

The remaining paper is organised as follows: problem formulation, modeling and proposed solution, numerical results and the conclusion are described in Sections \ref{sec_probformulation}-\ref{sec_conclusion}.

%%%%%%%%%%%%%%%%%%%%%%%%%%%%%%%%%%%%%%%%%%%%%%%%%%%
\section{Problem Formulation\label{sec_probformulation}}

The graphical abstracts for the two problems addressed in this paper are visualised in Fig. \ref{fig_prob1} and \ref{fig_prob2}.

\begin{problem}\label{problem1}
Given an aerial image find all sub-images similar to a given reference image representing a fixed-size object. An example application is the identification of all hay bales in an image.
\end{problem}

\begin{problem}\label{problem2}
Given an aerial image find all sub-images similar to a given reference image representing part of a variable-size image patch of interest. An example application is the identification of all patches representing damaged crop areas in an aerial image of an agricultural field.
\end{problem}

Input data to both problems are colored satellite or aerial images, $I\in \mathbb{I}_{[0,255]}^{N_x\times N_y \times 3}$, of dimension $N_x\times N_y$ and with RGB-channels for color. In general, the smaller-scale reference image, $i^\text{ref}\in \mathbb{I}_{[0,255]}^{N_x^\text{ref}\times N_y^\text{ref} \times 3}$, used for sub-image similarity search can be an arbitrary image. Throughout the experiments of this paper, the reference image is selected as a small sub-image of $I$. See Fig. \ref{fig_10data} and \ref{fig_10refimg} for illustration.

%%%%%%%%%%%%%%%%%%%%%%%%%%%%%%%%%%%%%%%%%%%%%%%%%%%
\section{Proposed Solution\label{sec_solnMethod}}

% ----------------Main Alg---------------------------
%\vspace{-0.1cm}
\begin{algorithm}
\SetKwInOut{Subfunctions}{\textbf{Subfunctions}}
\SetKwInOut{Input}{\textbf{Data Input}}
\SetKwInOut{Hyperparameters}{\textbf{Hyperparam.\hspace{0.05cm}}}
\SetKwInOut{Output}{\textbf{Final Result}}
\DontPrintSemicolon
\vspace{0.15cm}
\Subfunctions{$\mathcal{F}^\text{cost}(\cdot)$.}
\vspace{0.15cm}\hrule\vspace{0.15cm}
\Hyperparameters{$M>0$.}
\vspace{0.15cm}\hrule\vspace{0.15cm}
\Input{Images $I\in \mathbb{I}_{[0,255]}^{N_x\times N_y \times 3},~i^\text{ref}\in \mathbb{I}_{[0,255]}^{N_x^\text{ref}\times N_y^\text{ref} \times 3}$.}
\vspace{0.15cm}\hrule\vspace{0.15cm}
$X_\text{optl},Y_\text{optl},C_{\text{cost},\text{optl}} \leftarrow \{0\}_{0}^{M},\{0\}_{0}^{M},\{\infty\}_{0}^{M}$~{\color{gray}\%Initialisation.}\;
\For{$x\in \mathcal{X}_\text{space}$}
{
\For{$y\in \mathcal{Y}_\text{space}$}
{
$c_\text{cost} \leftarrow \mathcal{F}^\text{cost}(I,i^\text{ref},x,y,N_x^\text{ref},N_y^\text{ref})$.\;
\If{$c_\text{cost}<C_{\text{cost},\text{optl}}[M]$}
{
Insert $x,y,c_\text{cost}$ into ordered $X_\text{optl},Y_\text{optl},C_{\text{cost},\text{otpl}}$.\;
}
} % end \For{$y\in\matchcal{Y}_\text{space}$}
} % end \For{$x\in\matchcal{X}_\text{space}$}
\For{$\{X_\text{optl},Y_\text{optl}\}$}
{
Filter out such that only non-overlapping sub-images remain.\;
}
\vspace{0.15cm}\hrule\vspace{0.15cm}
\Output{$\{ X_\text{optl},Y_\text{optl} \}$.}
%
%\vspace{0.15cm}
%
\caption{Multi-Occurrences Sub-Image Search}\label{alg1}
\end{algorithm}
%\vspace{-0.5cm}
% ----------------------------------

Problem \ref{problem1} and \ref{problem2} differ in that either \emph{fixed-size} objects such as hay bales or \emph{variable-size} patches such as damaged crop areas in fields are searched, respectively. As will be shown, Problem \ref{problem2} is a generalisation of Problem \ref{problem1}. Both Problem \ref{problem1} and \ref{problem2} involve multiple-occurrences sub-image search. 

Algorithm \ref{alg1} describes the basic framework for multiple-occurrences sub-image search. This framework can be used for exhaustive search, where
\begin{equation}
\mathcal{X}_\text{space}=\{0,1,\dots,N_x\}, \quad \mathcal{Y}_\text{space}=\{0,1,\dots,N_y\}.
\label{eq_mathcalXY}
\end{equation}

In this paper, two ideas are proposed to (i) quantitatively achieve faster solve times, while (ii) qualitatively maintaining a similar performance to exhaustive search. Both ideas evolve around search space reduction and are presented in the following two subsections. The corresponding methods shall be denoted as \textsf{APTS-v1} and \textsf{APTS-v2}, since both are based on the APTS-algorithm (\cite{plessen2023posteriori}). 

\subsection{\textsf{APTS-v1}\label{subsec_APTSv1}}

\begin{figure}%https://tex.stackexchange.com/questions/165508/remove-a-b-from-subfigure-numbering-but-keep-the-subfigure-caption
\captionsetup[subfigure]{labelformat=empty}%<----If use this then NO '(a) Ex. 1' but just 'Ex. 1' (ie without the (a))
\centering
  \includegraphics[width=.9999\linewidth]{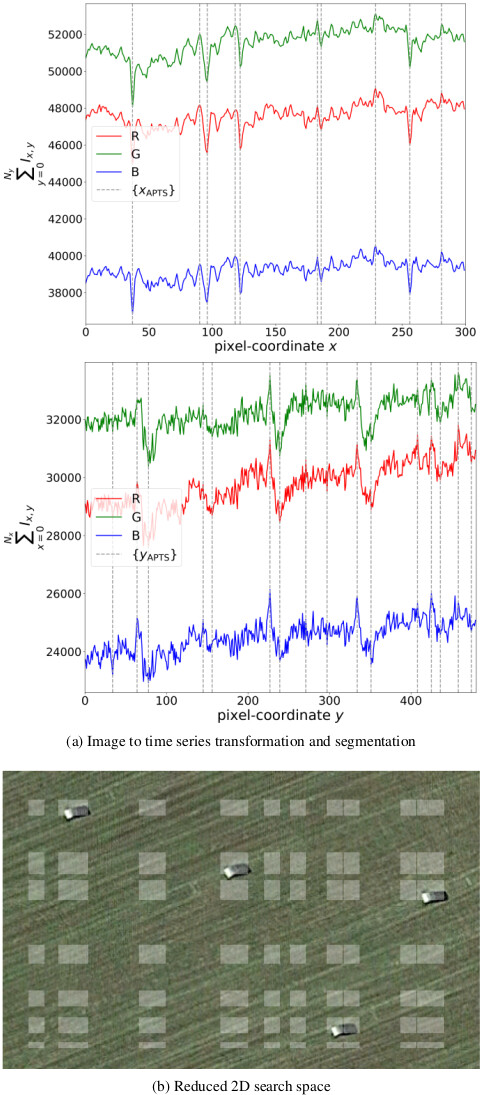}
\caption{(a) Virtual time series and their segmentations along $x$- and $y$-channels, respectively. Each channel consists of 3 time series, one for each RGB-color. (b) Reduced search space using the segmentation instances plus a tolerance as sampling points for sub-image search. The reduced search space shows the $(x,y)$-coordinates of the top-left corner of any sub-image of dimension $N_x^\text{ref}\times N_y^\text{ref}$ measured in similarity to the reference image.}
\label{fig_ex1_apts}
\end{figure}

%------------------

The following high-level algorithm is proposed to address Problem \ref{problem1} according to Fig. \ref{fig_prob1}:

\begin{itemize}
\item[1a.] Image to time series transformation: By summing along image rows and columns multivariate time series for the three RGB-channels are generated, $\{s_x\}_{x=0}^{N_x-1}$ and $\{s_y\}_{y=0}^{N_y-1}$ where $s_x\in \mathbb{I}_{[0,\infty)}^3$ and $s_y\in \mathbb{I}_{[0,\infty)}^3$, with 
\begin{subequations}
\begin{align}
s_x &= \sum_{y=0}^{N_y}I_{x,y},~x=0,\dots,N_x,\label{subeq_sx}\\
s_y &= \sum_{x=0}^{N_x}I_{x,y},~y=0,\dots,N_y.\label{subeq_sy}
\end{align}
\label{eq_timeseries}
\end{subequations}
 
\item[1b.] Time series segmentation: Both time series \eqref{subeq_sx} and \eqref{subeq_sy} are segmented according to the APTS-algorithm from \cite{plessen2023posteriori}. The set of resulting segmentation instances shall be denoted by $\{x_\text{APTS}\}$ and $\{y_\text{APTS}\}$, respectively. See Fig. \ref{fig_ex1_apts} (a) for illustration.

\item[2.] Based on $\{x_\text{APTS}\}$ and $\{y_\text{APTS}\}$ a reduced search space is built by shrinking $\mathcal{X}_\text{space}$ and $\mathcal{Y}_\text{space}$ from \eqref{eq_mathcalXY}. This is achieved by using $\{x_\text{APTS}\}$ and $\{y_\text{APTS}\}$ as main sampling points and adding a margin 
\begin{equation*}
m_\text{margin}=\text{max}\left( \left\lfloor{N_x^\text{ref}/p}\right\rfloor, \left\lfloor{N_y^\text{ref}/p}\right\rfloor \right),
\end{equation*}
around each of these sampling points, where $p>0$ is a hyperparameter (typically equal to either 1,2 or 4).

\item[3.] Apply Algorithm \ref{alg1} with reduced search spaces $\mathcal{X}_\text{space}$ and $\mathcal{Y}_\text{space}$.
\end{itemize}

\begin{table}%\begin{table*}
\vspace{0.3cm}
\centering
%\begin{small}
\begin{tabular}{|l|l|}
\hline %\hline
\rowcolor[gray]{0.8} Symbol & Value \\[2pt]%[-4pt]
\hline 
$\epsilon^\text{min}$  & 0.0001 \\
$\epsilon^\text{max}$  & 1 \\
$\gamma^\text{mult}$  & 2 \\
$\gamma^\text{close}$  & $\text{max}(0.01T,1)$ \\
\hline
\end{tabular}
%\end{small}
% ----------------
\caption{APTS-hyperparameters used uniformly throughout all experiments. Only hyperparameter $K^\text{max}$ (an upper bound on permitted segmentation instances) is varied throughout experiments with values according to Table \ref{tab_solve times}. $T>0$ denotes the time series length. For interpretation details see \cite{plessen2023posteriori}; for consistency the same notation and symbols are used.}
\label{tab_APTS_hp}
%\vspace{-0.1cm}
\end{table}%\end{table*}
  
Further details are discussed. First, the cost to evaluate the similarity between reference image and a sub-image block is calculated as,
\begin{equation}
\mathcal{F}^\text{cost}(I,i^\text{ref},x,y,N_x^\text{ref},N_y^\text{ref}) = \sum_{\bar{x}=0}^{N_x^\text{ref}}\sum_{\bar{y}=0}^{N_y^\text{ref}}\sum_{k=0}^{2} \left( I_{x+\bar{x},y+\bar{y},k} - i_{\bar{x},\bar{y},k}^\text{ref} \right)^2.
\label{eq_mathcalFcost}
\end{equation}
Second, a crucial characteristic of both Problem \ref{problem1} and \ref{problem2} is that a suitable number of sub-images similar to the reference image is not known a priori. This has two implications: (i) a hyperparameter $M>0$ is introduced as an upper bound on the desired number of similar sub-images to be found, (ii) sub-images must be \emph{ranked} in similarity such that only the $M$ most similar sub-images are returned after Steps 1-6 in Algorithm \ref{alg1}, before overlapping sub-images of those $M$ candidates are filtered out. In fact, the filtering Steps 7-8 in Algorithm \ref{alg1} result in practice in a much smaller number of identified sub-images than upper bound $M$. This has the additional benefit that fewer sub-images need to be clustered and processed in the TSP-solution step for patches contour generation. Especially for a stride step-size of 1 in Steps 2-3 of Algorithm \ref{alg1} a large $M$ has to be set to capture all relevant sub-images after the filtering step. The role of stride step-sizes larger than 1 will be discussed further below.   

Third, Table \ref{tab_APTS_hp} summarizes the APTS-hyperparameters. Hyperparameter $K^\text{max}$ from \cite{plessen2023posteriori}, an upper bound on the number of permitted segmentation instances, is absent from Table \ref{tab_APTS_hp} since this is varied for final experiments. All other hyperparameters are left constant throughout all experiments.

Finally to summarize, the scalar hyperparameters needed for \textsf{APTS-v1} are: $M>0$ from Algorithm \ref{alg1}, the hyperparameters from Table \ref{tab_APTS_hp}, $K^\text{max}>0$ and $p>0$.

\subsection{\textsf{APTS-v2}\label{subsec_APTSv2}}

% ----------------Apts-v2 Alg---------------------------
%\vspace{-0.1cm}
\begin{algorithm}
\SetKwInOut{Subfunctions}{\textbf{Subfunctions}}
\SetKwInOut{Input}{\textbf{Data Input}}
\SetKwInOut{Hyperparameters}{\textbf{Hyperparam.\hspace{0.05cm}}}
\SetKwInOut{Output}{\textbf{Final Result}}
\DontPrintSemicolon
\vspace{0.15cm}
\Subfunctions{$\mathcal{F}^\text{cost}(\cdot)$, $\mathcal{F}^{j_\text{axis}}(\cdot)$, 
$\mathcal{F}^\text{APTS}(\cdot)$.}
\vspace{0.15cm}\hrule\vspace{0.15cm}
\Hyperparameters{$\epsilon^\text{min}$, $\epsilon^\text{max}$, $\gamma^\text{mult}$, $\gamma^\text{close}$, $K^\text{max}$, $M$, $p$.}
\vspace{0.15cm}\hrule\vspace{0.15cm}
\Input{Images $I\in \mathbb{I}_{[0,255]}^{N_x\times N_y \times 3},~i^\text{ref}\in \mathbb{I}_{[0,255]}^{N_x^\text{ref}\times N_y^\text{ref} \times 3}$.}
\vspace{0.15cm}\hrule\vspace{0.15cm}
$X_\text{optl}^\text{both,axes},Y_\text{optl}^\text{both,axes} \leftarrow \{ \},\{ \}$.~{\color{gray}\%Initialisation.}\;
\For{$j_\text{axis} \in \{ 0,1\}$}
{
$i_{j_\text{axis}}^\text{ref} \leftarrow \mathcal{F}^{j_\text{axis}}(i^\text{ref},j_\text{axis},N_x^\text{ref},N_y^\text{ref})$~{\color{gray}\%Get 1D ref. img.}\;
Algorithm \ref{alg1} lines 1-6.\;
$X_\text{optl}^\text{both,axes},Y_\text{optl}^\text{both,axes}  \leftarrow X_\text{optl}^\text{both,axes}+X_\text{optl}, Y_\text{optl}^\text{both,axes}+Y_\text{optl} $.\;
} % end \For{$j_\text{axis} \in \{ 0,1\}$}
Algorithm \ref{alg1} lines 7-8 using $X_\text{optl}^\text{both,axes},Y_\text{optl}^\text{both,axes} $.\;
\vspace{0.15cm}\hrule\vspace{0.15cm}
\Output{$\{ X_\text{optl}^\text{both,axes},Y_\text{optl}^\text{both,axes} \}$.}
%
%\vspace{0.15cm}
%
\caption{\textsf{APTS-v2}}\label{alg2}
\end{algorithm}
%\vspace{-0.5cm}
% ----------------------------------

Steps 2-4 of basic Algorithm \ref{alg1} are of complexity order $\mathcal{O}(N_xN_yN_x^\text{ref}N_y^\text{ref}3)$.
By APTS-acceleration the complexity part $\mathcal{O}(N_xN_y)$ is in practice significantly reduced (quantitative results below in Section \ref{sec_IllustrativeEx}). To further reduce complexity, Algorithm \ref{alg2} is proposed. The corresponding complexity is $\mathcal{O}(N_x^\text{ref}+N_y^\text{ref}+N_xN_y(N_x^\text{ref}+3N_y^\text{ref}) + N_xN_y(N_y^\text{ref}+3N_x^\text{ref}))$. Instead of the full reference image, $i^\text{ref}\in \mathbb{I}_{[0,255]}^{N_x^\text{ref}\times N_y^\text{ref} \times 3}$, a smaller approximation is computed and used in Step 3 of Algorithm \ref{alg2}, where
\begin{equation*}
\mathcal{F}^{j_\text{axis}}(i^\text{ref},j_\text{axis},N_x^\text{ref},N_y^\text{ref})=\begin{cases} \left\{ \sum_{y=0}^{N_y^\text{ref}}i_{x,y}^\text{ref}\right\}_{x=0}^{N_x^\text{ref}}, & \text{if}~j_\text{axis}=0, \\ \left\{ \sum_{x=0}^{N_x^\text{ref}}i_{x,y}^\text{ref}\right\}_{y=0}^{N_y^\text{ref}}, & \text{if}~j_\text{axis}=1.
\end{cases}
\end{equation*}
The smaller reference image approximations along both axes are of dimensions $i_{0}^\text{ref}\in \mathbb{I}_{[0,\infty)}^{N_x^\text{ref}\times 1 \times 3}$ and $i_{1}^\text{ref}\in \mathbb{I}_{[0,\infty)}^{1\times N_y^\text{ref} \times 3}$. For each case $j_\text{axis} \in \{ 0,1\}$, Algorithm \ref{alg1} is then invoked according to Step 4 of Algorithm \ref{alg2}, before identified segmentation instances are merged in Step 5.

Several more comments are made. First, the simplifications are achieved by summing along the $x$- and $y$-axis, respectively. Notice that this \emph{still} maintains some spatial information over the reference image. A further and maximal simplification can be achieved by simultaneously summing along \emph{both} axes into a single scalar for each RGB-channel. This was also tested but found to be too inaccurate for practical applications.

Second, the set of scalar hyperparameters needed for \textsf{APTS-v2} is identical to the ones needed for \textsf{APTS-v1}.

\subsection{Clustering and Contour Generation\label{subsec_clustering}}

\begin{figure}%https://tex.stackexchange.com/questions/165508/remove-a-b-from-subfigure-numbering-but-keep-the-subfigure-caption
\captionsetup[subfigure]{labelformat=empty}%<----If use this then NO '(a) Ex. 1' but just 'Ex. 1' (ie without the (a))
\centering
  \includegraphics[width=.9999\linewidth]{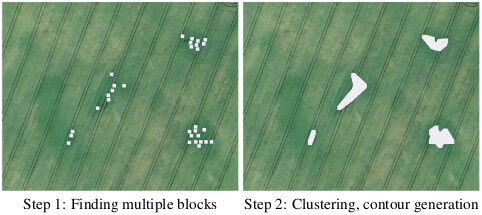}
\caption{Problem \ref{problem2}: After the identification of sub-images similar to the reference image these are further clustered before patches contours are determined by solving a traveling salesman problem.}
\label{fig_Ex9_step12}
\end{figure}

For Problem \ref{problem2} and the identification of variable-size patches additional processing is required. Once a set of sub-images similar to the reference image is identified, these are first clustered before contours for these clusters are then generated. For illustration, see Fig. \ref{fig_Ex9_step12}. In this paper, for the contour generation step a Traveling Salesman Problem (TSP) is solved using the heuristic method from \cite{plessen2024path}.

%%%%%%%%%%%%%%%%%%%%%%%%%%%%%%%%%%%%%%%%%%%%%%%%%%%%%%%%%%%%%%%%
\section{Numerical Results and Extending Discussion\label{sec_IllustrativeEx}}

\begin{figure}%https://tex.stackexchange.com/questions/165508/remove-a-b-from-subfigure-numbering-but-keep-the-subfigure-caption
\captionsetup[subfigure]{labelformat=empty}%<----If use this then NO '(a) Ex. 1' but just 'Ex. 1' (ie without the (a))
\centering
  \includegraphics[width=.99\linewidth]{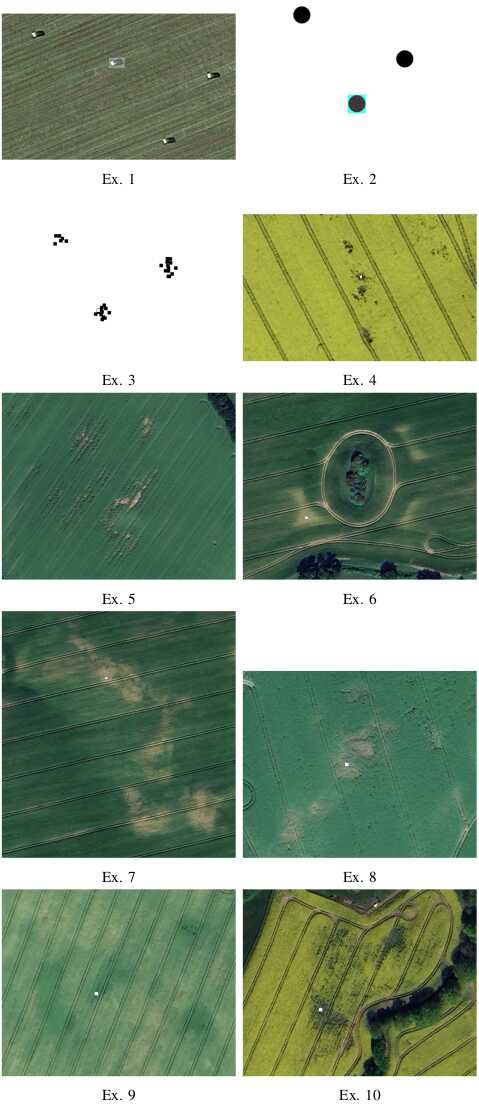}
\caption{Experiments: Problem data. 10 images with small rectangular reference images inside are displayed. For Ex. 1 and 2 reference images are indicated by the transparent rectangles. For the remaining experiments reference image locations are indicated by white rectangles for better clarity. Reference images are enlarged in Fig. \ref{fig_10refimg}.}
\label{fig_10data}
\end{figure}

\begin{figure}%https://tex.stackexchange.com/questions/165508/remove-a-b-from-subfigure-numbering-but-keep-the-subfigure-caption
\captionsetup[subfigure]{labelformat=empty}%<----If use this then NO '(a) Ex. 1' but just 'Ex. 1' (ie without the (a))
\centering
  \includegraphics[width=.99\linewidth]{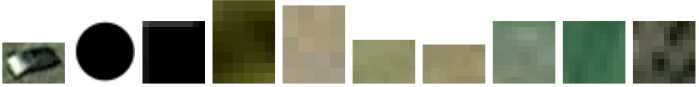}
\caption{Experiments: Problem data. Close-up of reference images in Fig. \ref{fig_10data}.}
\label{fig_10refimg}
\end{figure}

%tiny, scriptsize, footnotesize, small.

\begin{table}%\begin{table*}
\vspace{0.3cm}
\centering
%\begin{small}
\begin{tabular}{|c|r|r|r|}
\hline %\hline
\rowcolor[gray]{0.8} Ex. & \multicolumn{1}{c|}{METHOD 1} & \multicolumn{1}{c|}{METHOD 2} & \multicolumn{1}{c|}{METHOD 3} \\[-1pt]%[-4pt] 
\rowcolor[gray]{0.8} & \multicolumn{1}{c|}{(\textsf{Exhaustive})} & \multicolumn{1}{c|}{(\textsf{APTS-v1})} & \multicolumn{1}{c|}{(\textsf{APTS-v2})}\\[2pt]%[-4pt]
\hline 
%Index  & $M$ & $M,K^\text{max}^\text{APTS},p^\text{APTS}$ & $M,K^\text{max}^\text{APTS},p^\text{APTS}$ \\ 
Nr.  & $M$ & $M,K^\text{max},p$ & $M,K^\text{max},p$ \\ 
($N_x,N_y$)  & $\tau_\text{solve}$ & $\tau_\text{solve}$ & $\tau_\text{solve}$ \\ 
($N_x^\text{ref},N_y^\text{ref}$) & $\Delta \tau_\text{solve}$ & $\Delta \tau_\text{solve}$ & $\Delta \tau_\text{solve}$  \\[1pt]
\hline\hline
\textsf{1} & 10       & 10,20,4  & 15,20,4 \\
(300,480)  & 33.23s   & 6.11s    & 0.73s \\
(21,32)    &          & 82\%    & \textbf{-98\%} \\
\hline
\textsf{2} & 5        & 5,5,2    & 10,5,2 \\
(376,432)  & 45.06s   & 3.01s    & 0.66s \\
(35,34)      &          & 93\%    & \textbf{-99\%} \\
\hline
\textsf{3} & 1000      & 500,100,1    & 500,100,1 \\
(501,689)  & 16.20s   & 1.20s    & 0.49s \\
(10,10)      &          & -93\%    & \textbf{-97\%} \\
\hline
\textsf{4} & 1000      & 700,100,1    & 300,100,2 \\
(422,671)  & 4.86s    & 3.87s    & 1.12s \\
(8,6)      &          & -20\%    & \textbf{-77\%} \\
\hline
\textsf{5} & 2000      & 1000,100,1    & 1000,100,1 \\
(514,643)  & 4.83s    & 1.58s    & 1.51s \\
(5,4)      &          & -67\%    & \textbf{-69\%} \\
\hline
\textsf{6} & 3000      & 2000,100,2    & 2000,100,2 \\
(514,643)  & 10.54s   & 5.09s    & 3.27s \\
(7,10)      &          & -52\%    & \textbf{-69\%} \\
\hline
\textsf{7} & 10000      & 8000,100,1    & 4000,100,2 \\
(672,637)  & 26.85s   & 15.18s   & \textbf{5.84s} \\
(5,8)      &          & -43\%    & \textbf{-78\%} \\
\hline
\textsf{8} & 2000      & 3000,100,2    & 2500,100,2 \\
(514,643)  & 13.51s   & 8.65s   & 4.65s \\
(10,10)      &          & -36\%    & \textbf{-66\%} \\
\hline
\textsf{9} & 3000      & 1000,100,2    & 1200,100,2 \\
(514,643)  & 11.39s   & 4.04s    & 2.30s \\
(10,10)      &          & -65\%    & \textbf{-80\%} \\
\hline
\textsf{10} & 3000     & 2000,100,1    & 1200,100,1 \\
(514,643)   & 11.00s  & 6.55s    & 2.57s \\
(10,10)       &         & -40\%    & \textbf{-77\%} \\
\hline
\end{tabular}
%\end{small}
% ----------------
\caption{Experiments: Hyperparameter settings and quantitative evluation of three methods through statement of their solve times. For qualitative evaluation see Fig. \ref{fig_ex1to5} and \ref{fig_ex6to10}.}
\label{tab_solve times}
%\vspace{-0.1cm}
\end{table}%\end{table*}

\begin{table}%\begin{table*}
\vspace{0.3cm}
\centering
%\begin{small}
\begin{tabular}{|c|r|r|}
\hline %\hline
\rowcolor[gray]{0.8} Ex. 7 & \multicolumn{1}{c|}{(\textsf{Exhaustive})}  & \multicolumn{1}{c|}{(\textsf{APTS-v2})}\\%[2pt]%[-4pt]
\hline 
$\Delta_x,\Delta_y$  & $M$ & $M,K^\text{max},p$ \\ 
  & $\tau_\text{solve}$ & $\tau_\text{solve}$ \\[1pt]
\hline%\hline
4,4 & 900         & 100,100,4 \\
  & 0.70s     &  \textbf{0.12s}\\
  &     &  \textbf{-83\%}\\
\hline
\end{tabular}
%\end{small}
% ----------------
\caption{Experiment 7 (largest image dimensions out of all 10 experiments): When increasing stride step-sizes from from 1 to $(\Delta_x,\Delta_y)=(4,4)$ significant solve time reductions can be achieved. For \textsf{APTS-v2} the new solve time is 0.12s in comparison to the previous 5.84s in Table \ref{tab_solve times}. This is a reduction of -98\%.}
\label{tab_ex7}
%\vspace{-0.1cm}
\end{table}%\end{table*}

\begin{figure*}%https://tex.stackexchange.com/questions/165508/remove-a-b-from-subfigure-numbering-but-keep-the-subfigure-caption
\captionsetup[subfigure]{labelformat=empty}%<----If use this then NO '(a) Ex. 1' but just 'Ex. 1' (ie without the (a))
\centering
  \includegraphics[width=.99\linewidth]{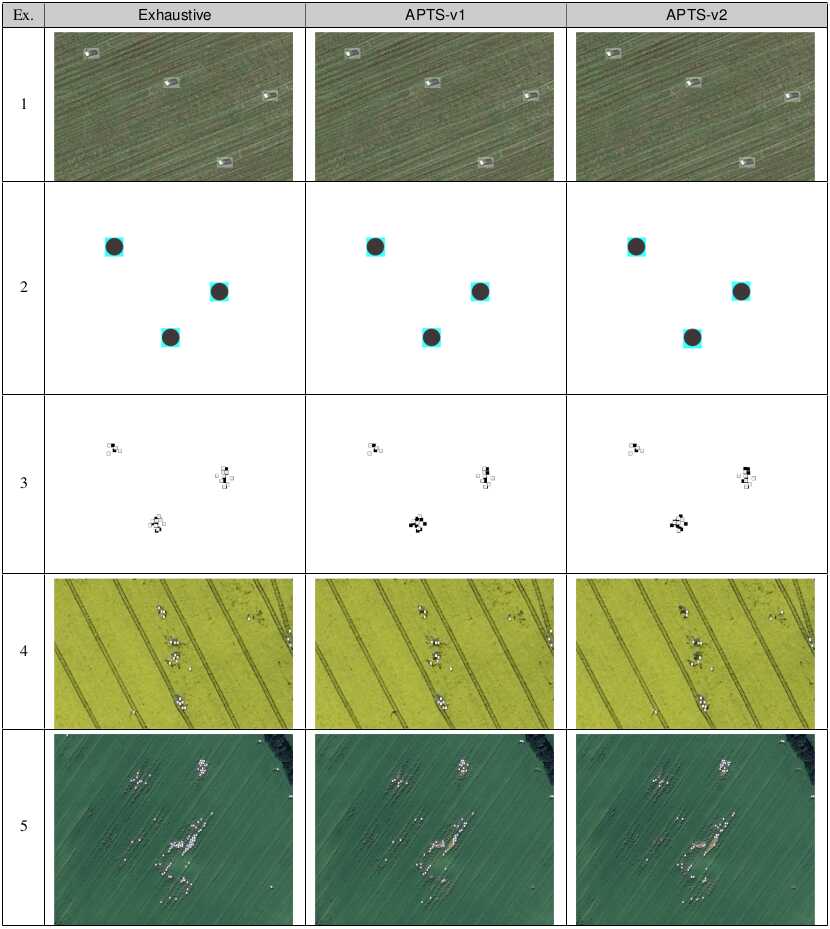}
\caption{Experiments: Qualitative evaluation of three methods for Ex. 1-5. See Table \ref{tab_solve times} for quantitative evaluation.}
\label{fig_ex1to5}
\end{figure*}

\begin{figure*}%https://tex.stackexchange.com/questions/165508/remove-a-b-from-subfigure-numbering-but-keep-the-subfigure-caption
\captionsetup[subfigure]{labelformat=empty}%<----If use this then NO '(a) Ex. 1' but just 'Ex. 1' (ie without the (a))
\centering
  \includegraphics[width=.99\linewidth]{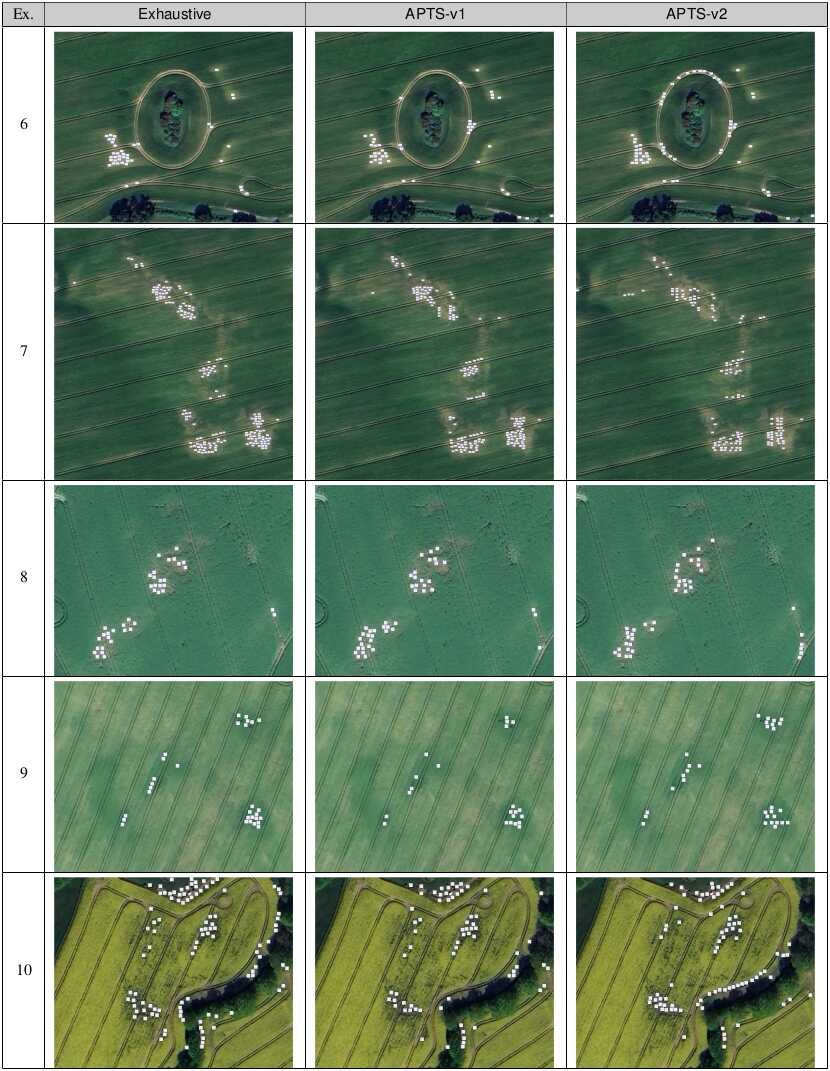}
\caption{Experiments: Qualitative evaluation of three methods for Ex. 6-10. See Table \ref{tab_solve times} for quantitative evaluation.}
\label{fig_ex6to10}
\end{figure*}

%---end OPT_2/2---

Proposed methods are evaluated on 10 diverse real-world and synthetic images. This is a small data set. However, this size permits to visually fully disclose qualitative performance within this paper. As mentioned before, a key characteristic of both Problem \ref{problem1} and \ref{problem2} is that a suitable number of sub-images similar to the reference image is not known a priori. Especially for Problem \ref{problem2} and variable-size patches identification a suitable number of patches is a subjective choice, making visual verification desirable. For this reason, experiments were conducted as follows. First, hyperparameter $M>0$ was tuned for the \textsf{Exhaustive} method until a visually satisfactory number of suitable sub-images was identified. Second, hyperparameters for \textsf{APTS-v1} and \textsf{APTS-v2} were then selected until a visually comparative performance was achieved. Qualitative results are summarised in Fig. \ref{fig_ex1to5} and \ref{fig_ex6to10}. Quantitative results are summarised in Table \ref{tab_solve times}.

In each experiment an image and a small reference-image part of the main image is provided as data input. See Fig. \ref{fig_10data} for illustration. The reference images are further enlarged in Fig. \ref{fig_10refimg}. Image sizes are stated in the left-most column of Table \ref{tab_solve times}. The main image has on average a size of 481 x 612 pixels. The reference image size on average is 12 x 13 pixels. All experiments were run on a laptop running Ubuntu 22.04 equipped with an Intel Core i9 CPU @5.50GHz×32 and 32 GB of memory. 

Multiple comments are made. First, \textsf{APTS-v2} quantitatively yields solve time reductions of between -66\% to -99\% or 2 orders of magnitude, while qualitatively yielding comparative visual results. The maximum solve time reduction of -99\% is achieved for the synthetic example Ex. 2. with noise-free white background and three black balls to be identified. This example provides basic proof-of-concept of proposed acceleration method. It further underlines that best performance is achieved for noise-free uniform backgrounds, thereby suggesting image pre-processing as a potential focus of future work.

Second, an important finding was that a single set of hyperparameters $(M,K^\text{max},p)$ that uniformly worked for all 10 experiments could \emph{not} be found. Crucially, this also holds for the \textsf{Exhaustive} method, where a single $M$ did also not suitably solve all experiments. This finding is not unexpected. Each experiment is a unique problem consisting of 1 main image and 1 reference image. In fact, the observation that different $M$-choices had to be set for \textsf{Exhaustive} to solve different problems supports the argument of using \textsf{APTS-v2} instead. This is because for practical applications for both methods one must iteratively adjust hyperparameters to the given problem data. However, in contrast to the \textsf{Exhaustive} method for \textsf{APTS-v2} this hyperparameters iteration can be conducted \emph{much faster} because of its much faster solve times.

Third, all results in Fig. \ref{fig_ex1to5}-\ref{fig_ex6to10} and Table \ref{tab_solve times} are obtained for stride step-size of 1 (maximal computational complexity). Hence, for the \textsf{Exhaustive} method, $\mathcal{X}_\text{space}$ and $\mathcal{Y}_\text{space}$ in Algorithm \ref{alg1} are as in \eqref{eq_mathcalXY}. To reduce computational complexity a larger stride may be selected. In general, however, this comes at the expense of reduced accuracy of sub-image search. For example, for strides $(\Delta_x,\Delta_y)=(4,4)$ reduced search spaces $\mathcal{X}_\text{space}=\{0,4,\dots\}$ and $\mathcal{Y}_\text{space}=\{0,4,\dots\}$ result. For the corresponding implementations of \textsf{APTS-v1} and \textsf{APTS-v2}, time series generation and segmentation remain as before, however, any sampling point candidate $x_\text{APTS}$ or $y_\text{APTS}$ of the reduced search space is only permitted if for a corresponding modulo operation it holds, $x_\text{APTS}\%\Delta_x=0$ or $y_\text{APTS}\%\Delta_y=0$, where $\%$ denotes the modulo operator.

\begin{figure}%https://tex.stackexchange.com/questions/165508/remove-a-b-from-subfigure-numbering-but-keep-the-subfigure-caption
\captionsetup[subfigure]{labelformat=empty}%<----If use this then NO '(a) Ex. 1' but just 'Ex. 1' (ie without the (a))
\centering
  \includegraphics[width=.99\linewidth]{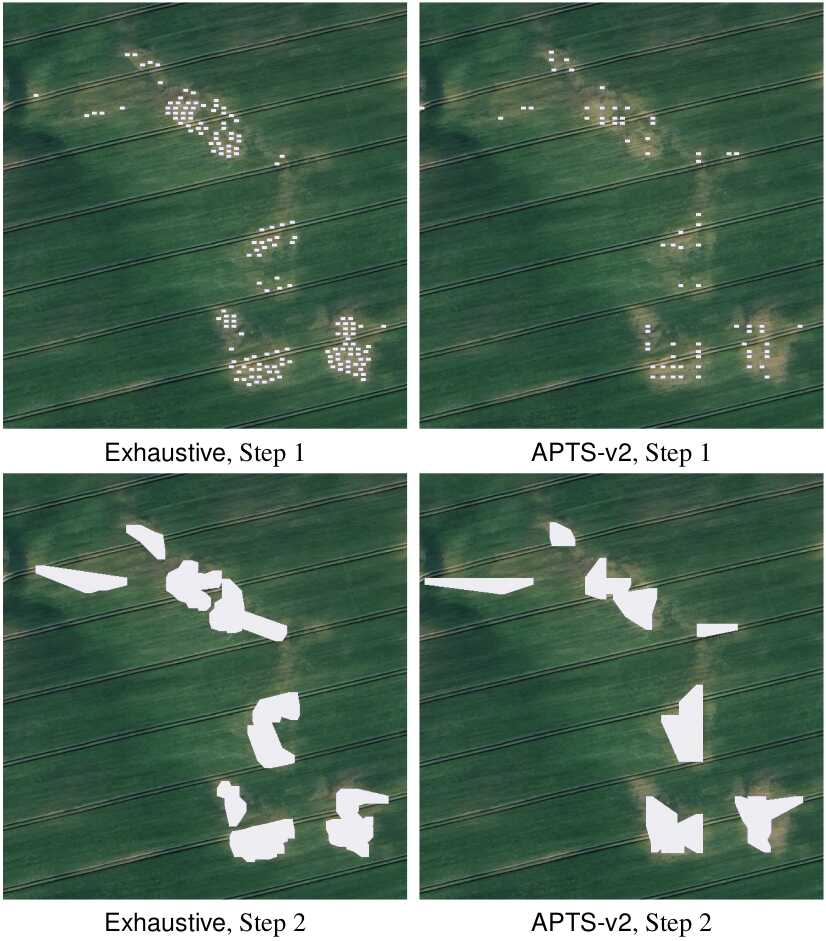}
\caption{Experiment 7: Qualitative evaluation of the results in Table \ref{tab_ex7} for increased stride step-sizes $(\Delta_x,\Delta_y)=(4,4)$. Step 1 and 2 illustrate results before and after clustering and contour generation for methods \textsf{Exhaustive} and \textsf{APTS-v2}, respectively.}
\label{fig_ex7_stepsize4}
\end{figure}

The largest image in Table \ref{tab_solve times} is Ex. 7, which also yields the slowest solve time of 5.84s for \textsf{APTS-v2}. When instead solving that example with stride step-sizes of $(\Delta_x,\Delta_y)=(4,4)$ then results in Table \ref{tab_ex7} and Fig. \ref{fig_ex7_stepsize4} are achieved. Two comments are made. First, hyperparameters had to be adjusted for both \textsf{Exhaustive} and \textsf{APTS-v2} to generate meaningful results. In particular, in both cases $M$ (the maximum permitted number of sub-image candidates) had to be reduced. Second and most importantly, \emph{much} faster solve times in comparison to the results for $(\Delta_x,\Delta_y)=(1,1)$ in Table \ref{tab_solve times} could be achieved. For \textsf{APTS-v2} the new solve time is 0.12s in comparison to the previous 5.84s. This is a reduction of -98\%. 

Finally, as Table \ref{tab_ex7} shows the solve time of 0.7s for the \textsf{Exhaustive} method is larger by a factor of 5.8 in comparison to the 0.12s of \textsf{APTS-v2}. This implies that in the same time almost 6 different reference-images could be processed, e.g., in the context of \emph{multi}-references search or classification applications.

%---------------------------------------
\subsection{Extending Discussion\label{subsec_extendingdiscussion}}

This subsection outlines other findings and aspects of presented method. First, above discussion focussed on \textsf{APTS-v2} rather than \textsf{APTS-v1} since it is the faster method. For reference images such as in Fig. \ref{fig_10refimg} with a regular or quasi-uniform texture \textsf{APTS-v2} is likely the best fit. However, for larger reference images with more variations in it, e.g. depicting specific objects, \textsf{APTS-v1} may be more suited since it can capture more spatial information since it does not apply the simplification step in Section \ref{subsec_APTSv2} characteristic for \textsf{APTS-v2}. Analysis for these settings is subject of ongoing work. 

Second, all of above experiments were also conducted for grayscale-versions of the images. On the one hand, grayscaled images reduce computational burden in Algorithm \ref{alg1} in that in \eqref{eq_mathcalFcost} it has not to be summed over the three RGB-channels anymore. On the other hand, a one-time conversion of the RGB-channels to grayscale is required. Here this conversion was implemented by weighted average of the RGB-channels, $I^\text{gray}_{x,y}=0.299I_{x,y,0} + 0.587I_{x,y,1} + 0.114I_{x,y,2},~\forall x=0,\dots,N_x,~\forall y=0,\dots,N_y$. The conversion is of complexity $\mathcal{O}(N_xN_y)$. Grayscale conversion reduced solve times for the \textsf{Exhaustive} method in all 10 experiments in comparison to the results in Table \ref{tab_solve times} for colored images. Surprisingly however, grayscale conversion reduced solve times for \textsf{APTS-v1} in only 8 cases and for \textsf{APTS-v2} in only 1 case. Thus, for \textsf{APTS-v2} in 9 out of 10 cases the grayscale conversion was more expensive than running the algorithm on the original colored image without conversion cost.

Third, a detail about images containing both an area of interest (e.g. the main field) and other area (e.g. part of a neighbouring field) is discussed. For Ex. 10 in Fig. \ref{fig_ex6to10} sub-images similar to the reference image are identified on the main field, but also outside of it. In practice, the latter can be filtered out by a geo-fencing algorithm. However, the most efficient method in case a contour of an area of interest is given, is to directly use that information in Step 2-3 of Algorithm \ref{alg1} for the design of $\mathcal{X}_\text{space}$ and $\mathcal{Y}_\text{space}$. This further accelerates solve times. 

Fourth, the method is suitable for parallelisation (and thus further acceleration), in particular, Step 2-6 of Algorithm \ref{alg1}.

Fifth, general benefits of proposed method are its simplicity and flexibility. It is library-free (e.g. no need for external solvers or routines), neural network-free (no need for labeled training data, architecture search and so forth), multi-scale in that it can work for arbitrary image dimensions, and does not assume special image pre-processing but instead directly works on the raw image. 

Finally, the multi-scale aspect is highlighted. In general and not discussed so far, application performance can greatly be steered by \emph{reference image choice}. A fast method such as \textsf{APTS-v2} that additionally can handle different image input dimensions is desirable. Then, one can quickly evaluate and focus on careful reference image selection for different applications.

%%%%%%%%%%%%%%%%%%%%%%%%%%%%%%%%%%%%%%%%%%%%%%%%%%%%%%%%%%%%%%%%%
\section{Conclusion\label{sec_conclusion}}

This paper addressed two image analysis tasks. First, the identification of fixed-size objects such as hay bales in an aerial image for a given reference image of the objects. Second, the identification of variable-size patches in an aerial image. Both tasks are related. The second differs in that identified sub-images similar to the reference image are further clustered before patches contours are determined by solving a traveling salesman problem. The underlying motivation for the second task is to serve as preparatory step for path planning between the identified patches for spot spraying with UAVs. One characteristic complexity of both tasks is that multiple sub-images similar to the reference image are searched, but a suitable number of similar sub-images is not known a priori. Instead, an upper bound has to be set as a hyperparameter.

The paper contributed by proposing a hierarchical method that accelerates sub-image search by conducting a transformation of the image to a multi-channel virtual time series, segmenting it, and using the segmentation instances as sampling points, thereby reducing the search space for sub-image search. Two algorithmic variations of different complexity, \textsf{APTS-v1} and \textsf{APTS-v2}, were presented, whereby the latter sums along rows and columns of the reference image to generate two reduced-size reference images to further minimize solve times.

Proposed methods were evaluated on 10 synthetic and real-world images and found to quantiatively deliver solve time reductions of up to 2 orders of magnitude, while qualitatively delivering comparative visual results to \textsf{Exhaustive} search.

For future work, instead of serving as preparatory step for path planning for spot spraying with UAVs, an area may also be surveyed to evaluate ripeness stadium of crops, before adapting logistics planning for biomass harvesting as in \cite{plessen2020gpu} or \cite{plessen2019coupling}.  

The second avenue of future work is to search for similarity with respect to multiple reference images instead of just one, and to further consider multiple instances of the same object in different scales and orientations.

%\section*{References} %---------------------------------------------------------------->UNCOMMENT FOR ArXiV!
%\bibliographystyle{model1a-num-names} %No.
%\bibliographystyle{model1-num-names} %No.
%\bibliographystyle{model2-names} %No.
%\bibliographystyle{model3-num-names} %No.
%\bibliographystyle{model4-names} %No.
\bibliographystyle{model5-names} %<----------
\bibliography{mybibfile.bib}

\begin{thebibliography}{20}
\expandafter\ifx\csname natexlab\endcsname\relax\def\natexlab#1{#1}\fi
\providecommand{\url}[1]{\texttt{#1}}
\providecommand{\href}[2]{#2}
\providecommand{\path}[1]{#1}
\providecommand{\DOIprefix}{doi:}
\providecommand{\ArXivprefix}{arXiv:}
\providecommand{\URLprefix}{URL: }
\providecommand{\Pubmedprefix}{pmid:}
\providecommand{\doi}[1]{\href{http://dx.doi.org/#1}{\path{#1}}}
\providecommand{\Pubmed}[1]{\href{pmid:#1}{\path{#1}}}
\providecommand{\bibinfo}[2]{#2}
\ifx\xfnm\relax \def\xfnm[#1]{\unskip,\space#1}\fi
%Type = Article
\bibitem[{Ahmad et~al.(1999)Ahmad, Kondo, Arima, Monta \&
  Mohri}]{ahmad1999weed}
\bibinfo{author}{Ahmad, U.}, \bibinfo{author}{Kondo, N.},
  \bibinfo{author}{Arima, S.}, \bibinfo{author}{Monta, M.}, \&
  \bibinfo{author}{Mohri, K.} (\bibinfo{year}{1999}).
\newblock \bibinfo{title}{Weed detection in lawn field using machine vision
  utilization of textural features in segmented area}.
\newblock {\it \bibinfo{journal}{Journal of the Japanese Society of
  Agricultural Machinery}\/},  {\it \bibinfo{volume}{61}\/},
  \bibinfo{pages}{61--69}.
%Type = Inproceedings
\bibitem[{Barrero et~al.(2016)Barrero, Rojas, Gonzalez \&
  Perdomo}]{barrero2016weed}
\bibinfo{author}{Barrero, O.}, \bibinfo{author}{Rojas, D.},
  \bibinfo{author}{Gonzalez, C.}, \& \bibinfo{author}{Perdomo, S.}
  (\bibinfo{year}{2016}).
\newblock \bibinfo{title}{Weed detection in rice fields using aerial images and
  neural networks}.
\newblock In {\it \bibinfo{booktitle}{2016 XXI Symposium on Signal Processing,
  Images and Artificial Vision}\/} (pp. \bibinfo{pages}{1--4}).
%Type = Inproceedings
\bibitem[{Eom \& Park(1990)}]{eom1990recognition}
\bibinfo{author}{Eom, K.-B.}, \& \bibinfo{author}{Park, J.}
  (\bibinfo{year}{1990}).
\newblock \bibinfo{title}{Recognition of shapes by statistical modeling of
  centroidal profile}.
\newblock In {\it \bibinfo{booktitle}{[1990] Proceedings of the 10th
  International Conference on Pattern Recognition}\/} (pp.
  \bibinfo{pages}{860--864}).
\newblock volume~\bibinfo{volume}{1}.
%Type = Article
\bibitem[{Ferro et~al.(2024)Ferro, S{\o}rensen \&
  Catania}]{ferro2024comparison}
\bibinfo{author}{Ferro, M.~V.}, \bibinfo{author}{S{\o}rensen, C.~G.}, \&
  \bibinfo{author}{Catania, P.} (\bibinfo{year}{2024}).
\newblock \bibinfo{title}{Comparison of different computer vision methods for
  vineyard canopy detection using uav multispectral images}.
\newblock {\it \bibinfo{journal}{Computers and Electronics in Agriculture}\/},
  {\it \bibinfo{volume}{225}\/}, \bibinfo{pages}{109277}.
%Type = Article
\bibitem[{Islam et~al.(2021)Islam, Rashid, Wibowo, Xu, Morshed, Wasimi, Moore
  \& Rahman}]{islam2021early}
\bibinfo{author}{Islam, N.}, \bibinfo{author}{Rashid, M.~M.},
  \bibinfo{author}{Wibowo, S.}, \bibinfo{author}{Xu, C.-Y.},
  \bibinfo{author}{Morshed, A.}, \bibinfo{author}{Wasimi, S.~A.},
  \bibinfo{author}{Moore, S.}, \& \bibinfo{author}{Rahman, S.~M.}
  (\bibinfo{year}{2021}).
\newblock \bibinfo{title}{Early weed detection using image processing and
  machine learning techniques in an australian chilli farm}.
\newblock {\it \bibinfo{journal}{Agriculture}\/},  {\it
  \bibinfo{volume}{11}\/}, \bibinfo{pages}{387}.
%Type = Inproceedings
\bibitem[{Ke et~al.(2004)Ke, Sukthankar, Huston, Ke \&
  Sukthankar}]{ke2004efficient}
\bibinfo{author}{Ke, Y.}, \bibinfo{author}{Sukthankar, R.},
  \bibinfo{author}{Huston, L.}, \bibinfo{author}{Ke, Y.}, \&
  \bibinfo{author}{Sukthankar, R.} (\bibinfo{year}{2004}).
\newblock \bibinfo{title}{Efficient near-duplicate detection and sub-image
  retrieval}.
\newblock In {\it \bibinfo{booktitle}{ACM multimedia}\/}
  (p.~\bibinfo{pages}{5}).
\newblock \bibinfo{organization}{Citeseer} volume~\bibinfo{volume}{4}.
%Type = Inproceedings
\bibitem[{Keogh et~al.(2006)Keogh, Wei, Xi, Lee \& Vlachos}]{keogh2006lb_keogh}
\bibinfo{author}{Keogh, E.}, \bibinfo{author}{Wei, L.}, \bibinfo{author}{Xi,
  X.}, \bibinfo{author}{Lee, S.-H.}, \& \bibinfo{author}{Vlachos, M.}
  (\bibinfo{year}{2006}).
\newblock \bibinfo{title}{Lb{\_}keogh supports exact indexing of shapes under
  rotation invariance with arbitrary representations and distance measures}.
\newblock In {\it \bibinfo{booktitle}{Proceedings of the 32nd International
  Conference on Very Large Data Bases}\/} (pp. \bibinfo{pages}{882--893}).
%Type = Article
\bibitem[{Li et~al.(2023)Li, Sheng, Zhang \& Zhang}]{li2023coverage}
\bibinfo{author}{Li, J.}, \bibinfo{author}{Sheng, H.}, \bibinfo{author}{Zhang,
  J.}, \& \bibinfo{author}{Zhang, H.} (\bibinfo{year}{2023}).
\newblock \bibinfo{title}{Coverage path planning method for agricultural
  spraying uav in arbitrary polygon area}.
\newblock {\it \bibinfo{journal}{Aerospace}\/},  {\it \bibinfo{volume}{10}\/},
  \bibinfo{pages}{755}.
%Type = Article
\bibitem[{Ospina \& Noguchi(2019)}]{ospina2019simultaneous}
\bibinfo{author}{Ospina, R.}, \& \bibinfo{author}{Noguchi, N.}
  (\bibinfo{year}{2019}).
\newblock \bibinfo{title}{Simultaneous mapping and crop row detection by fusing
  data from wide angle and telephoto images}.
\newblock {\it \bibinfo{journal}{Computers and Electronics in Agriculture}\/},
  {\it \bibinfo{volume}{162}\/}, \bibinfo{pages}{602--612}.
%Type = Article
\bibitem[{Plessen(2024)}]{plessen2024path}
\bibinfo{author}{Plessen, M.} (\bibinfo{year}{2024}).
\newblock \bibinfo{title}{Path planning for spot spraying with uavs combining
  tsp and area coverages}.
\newblock {\it \bibinfo{journal}{arXiv preprint arXiv:2408.08001}\/}, .
%Type = Article
\bibitem[{Plessen(2019)}]{plessen2019coupling}
\bibinfo{author}{Plessen, M.~G.} (\bibinfo{year}{2019}).
\newblock \bibinfo{title}{Coupling of crop assignment and vehicle routing for
  harvest planning in agriculture}.
\newblock {\it \bibinfo{journal}{Artificial Intelligence in Agriculture}\/},
  {\it \bibinfo{volume}{2}\/}, \bibinfo{pages}{99--109}.
%Type = Article
\bibitem[{Plessen(2020{\natexlab{a}})}]{plessen2020gpu}
\bibinfo{author}{Plessen, M.~G.} (\bibinfo{year}{2020}{\natexlab{a}}).
\newblock \bibinfo{title}{{GPU}-accelerated logistics optimisation for biomass
  production with multiple simultaneous harvesters tours, fields and plants}.
\newblock {\it \bibinfo{journal}{Biomass and Bioenergy}\/},  {\it
  \bibinfo{volume}{141}\/}, \bibinfo{pages}{105650}.
%Type = Inproceedings
\bibitem[{Plessen(2020{\natexlab{b}})}]{plessen2020integrated}
\bibinfo{author}{Plessen, M.~G.} (\bibinfo{year}{2020}{\natexlab{b}}).
\newblock \bibinfo{title}{Integrated time series summarization and prediction
  algorithm and its application to covid-19 data mining}.
\newblock In {\it \bibinfo{booktitle}{2020 IEEE International Conference on Big
  Data}\/} (pp. \bibinfo{pages}{4945--4954}).
%Type = Inproceedings
\bibitem[{Plessen(2023)}]{plessen2023posteriori}
\bibinfo{author}{Plessen, M.~G.} (\bibinfo{year}{2023}).
\newblock \bibinfo{title}{A posteriori trading-inspired model-free time series
  segmentation}.
\newblock In {\it \bibinfo{booktitle}{2023 IEEE International Conference on Big
  Data}\/} (pp. \bibinfo{pages}{5888--5896}).
%Type = Article
\bibitem[{Reedha et~al.(2022)Reedha, Dericquebourg, Canals \&
  Hafiane}]{reedha2022transformer}
\bibinfo{author}{Reedha, R.}, \bibinfo{author}{Dericquebourg, E.},
  \bibinfo{author}{Canals, R.}, \& \bibinfo{author}{Hafiane, A.}
  (\bibinfo{year}{2022}).
\newblock \bibinfo{title}{Transformer neural network for weed and crop
  classification of high resolution uav images}.
\newblock {\it \bibinfo{journal}{Remote Sensing}\/},  {\it
  \bibinfo{volume}{14}\/}, \bibinfo{pages}{592}.
%Type = Inproceedings
\bibitem[{Sebe et~al.(1999)Sebe, Lew \& Huijsmans}]{sebe1999multi}
\bibinfo{author}{Sebe, N.}, \bibinfo{author}{Lew, M.~S.}, \&
  \bibinfo{author}{Huijsmans, D.~P.} (\bibinfo{year}{1999}).
\newblock \bibinfo{title}{Multi-scale sub-image search}.
\newblock In {\it \bibinfo{booktitle}{Proceedings of the Seventh ACM
  International Conference on Multimedia (Part 2)}\/} (pp.
  \bibinfo{pages}{79--82}).
%Type = Article
\bibitem[{S{\o}gaard \& Olsen(2003)}]{sogaard2003determination}
\bibinfo{author}{S{\o}gaard, H.~T.}, \& \bibinfo{author}{Olsen, H.~J.}
  (\bibinfo{year}{2003}).
\newblock \bibinfo{title}{Determination of crop rows by image analysis without
  segmentation}.
\newblock {\it \bibinfo{journal}{Computers and Electronics in Agriculture}\/},
  {\it \bibinfo{volume}{38}\/}, \bibinfo{pages}{141--158}.
%Type = Article
\bibitem[{Tang et~al.(2016)Tang, Chen, Miao \& Wang}]{tang2016weed}
\bibinfo{author}{Tang, J.-L.}, \bibinfo{author}{Chen, X.-Q.},
  \bibinfo{author}{Miao, R.-H.}, \& \bibinfo{author}{Wang, D.}
  (\bibinfo{year}{2016}).
\newblock \bibinfo{title}{Weed detection using image processing under different
  illumination for site-specific areas spraying}.
\newblock {\it \bibinfo{journal}{Computers and Electronics in Agriculture}\/},
  {\it \bibinfo{volume}{122}\/}, \bibinfo{pages}{103--111}.
%Type = Article
\bibitem[{Veeranampalayam~Sivakumar et~al.(2020)Veeranampalayam~Sivakumar, Li,
  Scott, Psota, J.~Jhala, Luck \& Shi}]{veeranampalayam2020comparison}
\bibinfo{author}{Veeranampalayam~Sivakumar, A.~N.}, \bibinfo{author}{Li, J.},
  \bibinfo{author}{Scott, S.}, \bibinfo{author}{Psota, E.},
  \bibinfo{author}{J.~Jhala, A.}, \bibinfo{author}{Luck, J.~D.}, \&
  \bibinfo{author}{Shi, Y.} (\bibinfo{year}{2020}).
\newblock \bibinfo{title}{Comparison of object detection and patch-based
  classification deep learning models on mid-to late-season weed detection in
  uav imagery}.
\newblock {\it \bibinfo{journal}{Remote Sensing}\/},  {\it
  \bibinfo{volume}{12}\/}, \bibinfo{pages}{2136}.
%Type = Article
\bibitem[{Zhang \& Lu(2004)}]{zhang2004review}
\bibinfo{author}{Zhang, D.}, \& \bibinfo{author}{Lu, G.}
  (\bibinfo{year}{2004}).
\newblock \bibinfo{title}{Review of shape representation and description
  techniques}.
\newblock {\it \bibinfo{journal}{Pattern Recognition}\/},  {\it
  \bibinfo{volume}{37}\/}, \bibinfo{pages}{1--19}.

\end{thebibliography}
\nocite{*}

\end{document}